\documentclass[nohyperref]{article}
\usepackage{multirow}
\usepackage{microtype}
\usepackage{graphicx}
\usepackage{subfigure}
\usepackage{booktabs} 
\usepackage{stfloats}
\makeatletter
\newcommand\figcaption{\def\@captype{figure}\caption}
\newcommand\tabcaption{\def\@captype{table}\caption}
\usepackage{hyperref}

\usepackage[accepted]{icml2022}

\usepackage{amsmath}
\usepackage{amssymb}
\usepackage{mathtools}
\usepackage{amsthm}

\usepackage[capitalize,noabbrev]{cleveref}

\theoremstyle{plain}

\theoremstyle{definition}

\theoremstyle{remark}

\usepackage[textsize=tiny]{todonotes}

\icmltitlerunning{Protum: A New Method For Prompt Tuning Based on "[MASK]"}

\begin{document}

\twocolumn[
\icmltitle{Protum: A New Method For Prompt Tuning Based on "[MASK]"}



\icmlsetsymbol{equal}{*}

\begin{icmlauthorlist}
\icmlauthor{Pan He}{aff1}
\icmlauthor{Yuxi Chen}{aff1}
\icmlauthor{Yan Wang}{aff1}
\icmlauthor{Yanru Zhang}{aff1,aff2}
\end{icmlauthorlist}

\icmlaffiliation{aff1}{University of Electronic Science and Technology of China, Chengdu, China}
\icmlaffiliation{aff2}{Shenzhen Institute for Advanced Study, UESTC, China}

\icmlcorrespondingauthor{Yanru Zhang}{yanruzhang@uestc.edu.cn}

\icmlkeywords{Prompt Tuning, Pre-training, ResNet}

\vskip 0.3in
]



\printAffiliationsAndNotice{\icmlEqualContribution} 

\begin{abstract}
Recently, prompt tuning \cite{lester2021power} has gradually become a new paradigm for NLP, which only depends on the representation of the words by freezing the parameters of pre-trained language models (PLMs) to obtain remarkable performance on downstream tasks. It maintains the consistency of Masked Language Model (MLM) \cite{devlin2018bert} task in the process of pre-training, and avoids some issues that may happened during fine-tuning. Naturally, we consider that the "[MASK]" tokens carry more useful information than other tokens because the model combines with context to predict the masked tokens. Among the current prompt tuning methods, there will be a serious problem of random composition of the answer tokens in prediction when they predict multiple words so that they have to map tokens to labels with the help verbalizer. In response to the above issue, we propose a new \textbf{Pro}mpt \textbf{Tu}ning based on "[\textbf{M}ASK]" (\textbf{Protum}) method in this paper, which constructs a classification task through the information carried by the hidden layer of "[MASK]" tokens and then predicts the labels directly rather than the answer tokens. At the same time, we explore how different hidden layers under "[MASK]" impact on our classification model on many different data sets. Finally, we find that our \textbf{Protum} can achieve much better performance than fine-tuning after continuous pre-training with less time consumption. Our model facilitates the practical application of large models in NLP.
\end{abstract}

\section{Introduction}

As the proposal of a series of pre-trained models (PLMs) based on large-scale corpus, the benchmark of downstream tasks in NLP has been pushed to a new level and fine-tuning (FT) on PLMs has gradually become the main paradigm of Natural Language Understanding (NLU) and Natural Language generation (NLG).
Many scholars have also made some decent improvements in both pre-training and fine-tuning, which make the performance of the model surpassing human beings in a few NLU tasks.
Unfortunately, better results also come at the unbearable consumption of time and computing power.
At present, the pre-trained models are based on transformer \cite{vaswani2017attention} , in which NLG (GPT3 \cite{brown2020language}, T5 \cite{raffel2019exploring}, CPT \cite{shao2021cpt}, etc.) uses both the encoder and decoder of  transformer, while NLU (BERT \cite{devlin2018bert}, RoBERTa \cite{liu2019roberta}, ALBERT \cite{lan2019albert}, etc.) only uses the encoder to obtain the representation of sentences.
What's more, the time complexity of attention mechanism of transformer \cite{vaswani2017attention} is $O(n^2)$. The longer the sentence whether in pre-training or fine-tuning, the higher time consumption, which poses a great obstacle to their application.
At the same time, as the tremendous volume of corpora, the size of models become larger and larger with billions of parameters. Not only does it consume a lot of time but is very easy to overfit when we fine-tune the large model on the small data set.
A constructive idea which named prompt tuning (PT) \cite{lester2021power} converts downstream tasks to cloze tasks by constructing templates artificially or automatically, which aims to predict the masked tokens by MLM, and then constructs an unique verbalizer from tokens to labels for a specific downstream task. In an example of classification for cinecisms, we construct a template first "$[X]$. Overall, this is a $[Z]$ movie."  by adding the prompt tokens. Secondly the MLM of the PLMs which fixed parameters predicts $[Z]$ as well as the process of pre-training, where $[X]$ represents a sentence in the cinecisms,$[Z]\in \{{\rm good, bad}\}$ and  $label = {\rm Verbalizer}([Z]) \in \{0,1\} $.
PT maintains the consistency between pre-training and prompt tuning, which can avoid some kinds of issues such as knowledge forgetting to a certain extent. Additionally , PT has very fast training speed because it only updates a small part of parameters except those of the PLMs.

\begin{figure}[h]
    \centering
    \includegraphics[scale = 0.75]{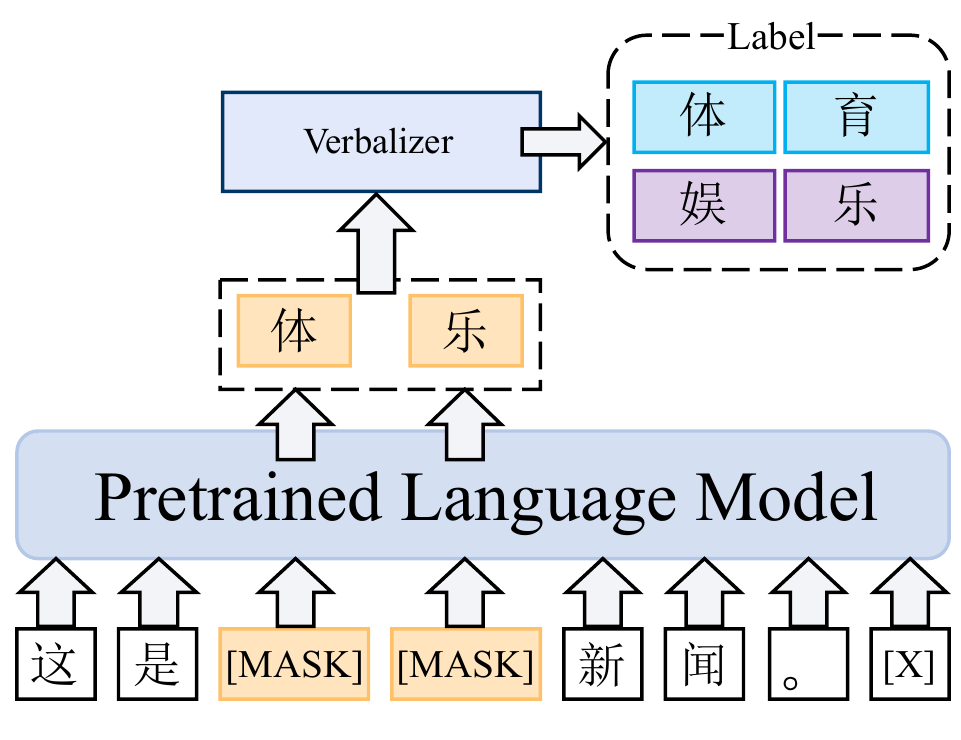}
    \caption{An instance of unreasonable tokens prediction for the multiple answer words in the News classification task of Chinese. $[X]$ represents a sentence of a News.}
    \label{fig:multi_words}
\end{figure}

However, influenced by different prompts, there still exit two main problems in the PT methods.
One is the influence of the diverse choices of $[Z]$ in prompt. In the cinecism classification task, $[Z]$ is not unique. It only needs to indicate that both positive and negative adjectives are acceptable, such as "great" \& "worse", "awesome" \& "terrible, "fantastic" \& "awful", "amazing" \& "sucked" and so on. Somehow, whether it is constructed manually or automatically, it means that only two selected words are predicted from vocabulary, and other group words cannot be predicted. Theoretically, the answer tokens of another groups are also correct. In our opinion, the information in many PLMs is not fully utilized and this factor has a great impact on the effectiveness of PT.
The second fatal issue is that the answer token $[Z]$ they construct is almost a single word, there will be a very serious problem of random tokens composition in MLM in the case of multiple words prediction. It is quite normal for classification tasks in those languages such as Chinese which can not use a single word to represent a category in most cases.
This phenomenon is shown in Figure \ref{fig:multi_words}.
Inspired by MLM, we believe "[MASK]" should contain more direct and useful information owing to combination with the context of "[MASK]" to predict the masked tokens. In this paper, we convert the prediction of tokens during PT to a classification task based on "[MASK]", which solves the problem that the current PT methods can not handle the issue of the multiple tokens prediction well.

This paper is organized as follows. The second part of our paper give certainly descriptions of current PT methods and their limitations. 
In section 3, firstly, we introduce \textbf{Protum-base} in detail and how we solve the issues above. Secondly, according to the ordinary results obtained by \textbf{Protum-base}, we propose \textbf{Protum-Res} on the basis of \textbf{Protum-base} for improving the performance on both large models and small models.
In section 4, we explore the effectiveness of our \textbf{Protum} in several NLP tasks, and then give explanations about the results. Finally, we conclude with a brief discussion of our research and expectation.
\section{Related Work}
Pre-training language models like BERT \cite{devlin2018bert}, RoBERTa \cite{liu2019roberta}, GPT \cite{radford2018improving} have been widely used for many natural language processing (NLP) tasks for several years. Although the language models perform well after being fine-tuned to adapt to series of specific downstream tasks, it needs to introduce and store a considerable amount of parameters, thus leading to a large cost on computation and memorizing. 
In order to solve this issue, many scholars have made a lot of efforts to research various useful approaches to it. In 2019, \cite {houlsby2019parameter} have proposed a kind of special unit which named adapter. By inserting the adapter between the layers of transformer, the decent results can be obtained by only updating the parameters of the adapters. In 2021, Google presents prompt tuning \cite{lester2021power} and converts various the downstream tasks of FT to the tasks of PT with the capacity of MLM to predict the masked tokens. Besides, they also make sufficient researches about how the length of prompts and the scale of the parameters of the PLMs impact on the performance of several tasks. The conclusion shows that the longer the prompts or the larger the PLMs bring more satisfying results.

According to the strategies of prompt-designing, the prompt-based methods can be roughly split into two types: discrete prompts and continuous prompts. Discrete prompts, also known as hard prompts, generate templates composed of natural language strings. Continuous prompts, though, directly work on the continuous embedding space rather than find the exact natural language tokens, so that the models can be optimized through back-propagation. Owing to the prior information provided by the hard prompts or soft prompts, researchers quickly realize the huge potentiality on the few-shot and the zero-shot learning.
In this year, both Prefix-tuning \cite{li2021prefix} and P-tuning \cite{liu2021gpt} are proposed, which are the methods of only updating the parameters of the additional layers of continuous prompts to improve the performance on the downstream tasks comparing with the PT methods before.
P-tuning also designs the experiments to figure out the effectiveness of their methods on few-shot data sets and the results indicate that PT is better than FT on the few-shot learning. \cite{gu2021ppt, schick2020s,zhang2021differentiable} also put forward many useful strategies to make both small and large language models become better few-shot learner.

In all of the PT methods mentioned, the proper verbalizers are selected to transfer the answer tokens predicted by MLM to labels. The disadvantage of the verbalizers is showing in the Figure \ref{fig:multi_words}. In the experiments of P-tuning v2 \cite{liu2021pv2}, the CLS head of the PLMs is taken to make classification for comparing with the verbalizers. However, the results indicate that the performances obtained by these two approaches are basically the same and the CLS head has not bring any improvements.

\section{Methodology}
In this section, we present our \textbf{Protum-base} and  \textbf{Protum-Res} simultaneously that a brand new way of prompt tuning based on "[MASK]" without any verbalizers and construction of models for various NLP downstream tasks.
\footnote{The code of our two types PT methods including \textbf{Protum-base} and \textbf{Protum-Res} will be released at two weeks later after refining.}
\subsection{Protum-base}
\begin{figure*}[htp]
    \centering
    \includegraphics[scale=0.7]{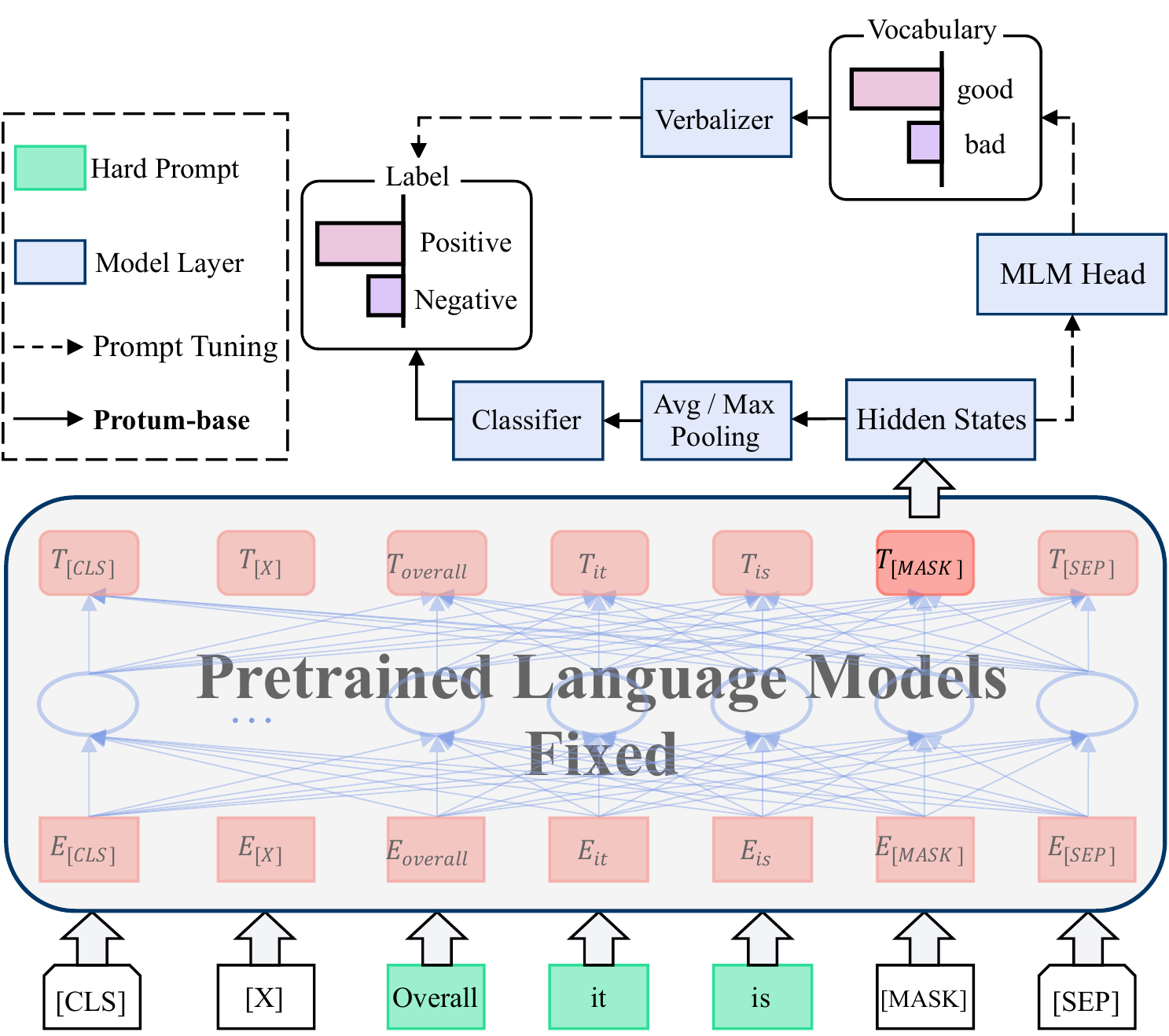}
    \caption{The difference of structure between Prompt tuning and our \textbf{Protum-base}. We replace the verbalizer with a classifier to solve the issue of random combination between the answers tokens. Besides, we change the way of soft prompt with parameters updating to a classifier.}
    \label{fig:protum_base}
\end{figure*}

Comparing with others PT methods, in our \textbf{Protum-base}, we take the hidden states of the masked tokens rather than MLM head or classification linear head to make classification while we get rid of the verbalizer. The structure of our \textbf{Protum-base} is shown in the Figure \ref{fig:protum_base}. In the processing of training MLM, with the context around the masked tokens, PLMs can infer what the correct tokens under the "[MASK]" are. Naturally, it is reasonable for classification to take the hidden states under the masked tokens into consideration directly rather than converting the hidden states to the specific answer tokens. The advantage of this strategy is that we do not need to care about what the masked words are while the classification layer connected to the PLMs will divide the hidden states of the masked tokens into the labels of a downstream task. Given a data set of a classification task $(X_i,Y_i)$, where $X_i$ represents a single sentence or sentences-pair and $Y_i$ is the label of $X_i$. And then we construct a PT data set $(T_i,Y_i)$ by adding the appropriate hard prompt tokens. We define $H_{T_i}$ as that the hidden states of the masked answer tokens extracted from the PLMs with fixed parameters such as BERT after inputting $T_i$.
\begin{equation}
    H_{T_i}={\rm PLMs}(T_i),H_{T_i} \in R^{N * W * M}
\end{equation}
where N and M represent the number of hidden layers and the hidden size of the PLMs simultaneously. Besides, W is the length of the masked tokens in $T_i$. 
\begin{equation}
    \tilde{H}_{T_i} = {\rm Max/Avg}(H_{T_i},{\rm dim}=2), \tilde{H}_{T_i}\in R^{N * M}
\end{equation}
\begin{equation}
    \hat{y_i} = {\rm Classifier}(\tilde{H}_{T_i}[j]), j\in [1,N], j\in Z
\end{equation}
Then $\tilde{H}_{T_i}$ extracted from the PLMs is taking to make max pool or average pool in the dimension of the length of the masked answer tokens. During the processing of pooling, no matter how many answer tokens there are, this step can handle the issue of random combination of the answer tokens when those PT methods predict multiple words. Lastly, $\tilde{H}_{T_i}[j]$ is the hidden states after pooling of the layer $j$ of the PLMs which are regard as the input for a learnable classifier to predict the labels. The whole process of our \textbf{Protum-base} is very simple which is a task of classifying the features of the text extracted from the PLMs in which the dimension of the features is hidden size. The MLM head \cite{han2021ptr} are regard as the input for classifier layers in the previous PT methods. It means that the hidden states of the masked token where their dimension are hidden size of the PLMs are converted to the MLM head with the vocabulary size dimension. 
\begin{equation}
    {\rm MLM}_i = {\rm MLMHead}(H_{T_i}[-1]), {\rm MLM}_i \in R^{W * V}
\end{equation}
\begin{equation}
    \hat{y_i} = {\rm Verbalizer}({\rm MLM}_i)
\end{equation}
where $H_{T_i}[-1]$ is the sequence output of $H_{T_i}$ and $V$ represents the size of vocabulary of the PLMs. In other words, they convert all downstream classification tasks of NLP to a specific classification task where the size of categories is $V$ but a few of them which mean the answer tokens for the tasks are useful. Obviously, those PT methods using MLM head weaken the performance and robustness of the PLMs.
\subsection{Protum-Res}
In the above chapter, the \textbf{Protum-base} which makes a better choice that it updates the parameters of a classifier rather than the soft prompt during PT.
Nevertheless, there still exists two serious issues in our model.
\begin{itemize}
    \item The parameters of the classifier are too small to satisfy the condition of the convergence for our model.
    \item Only the hidden states of one layer of the PLMs are brought to make classification. 
\end{itemize}
Obviously, it is not well making full usage for the hidden states of each layer so that this approach reduce the performance of models to some extend. Inspired by ResNet \cite{he2016resnet}, we introduce the \textbf{Protum-Res} based on the \textbf{Protum-base} we mentioned before. 

\begin{figure}
    \centering
    \includegraphics[scale=0.6]{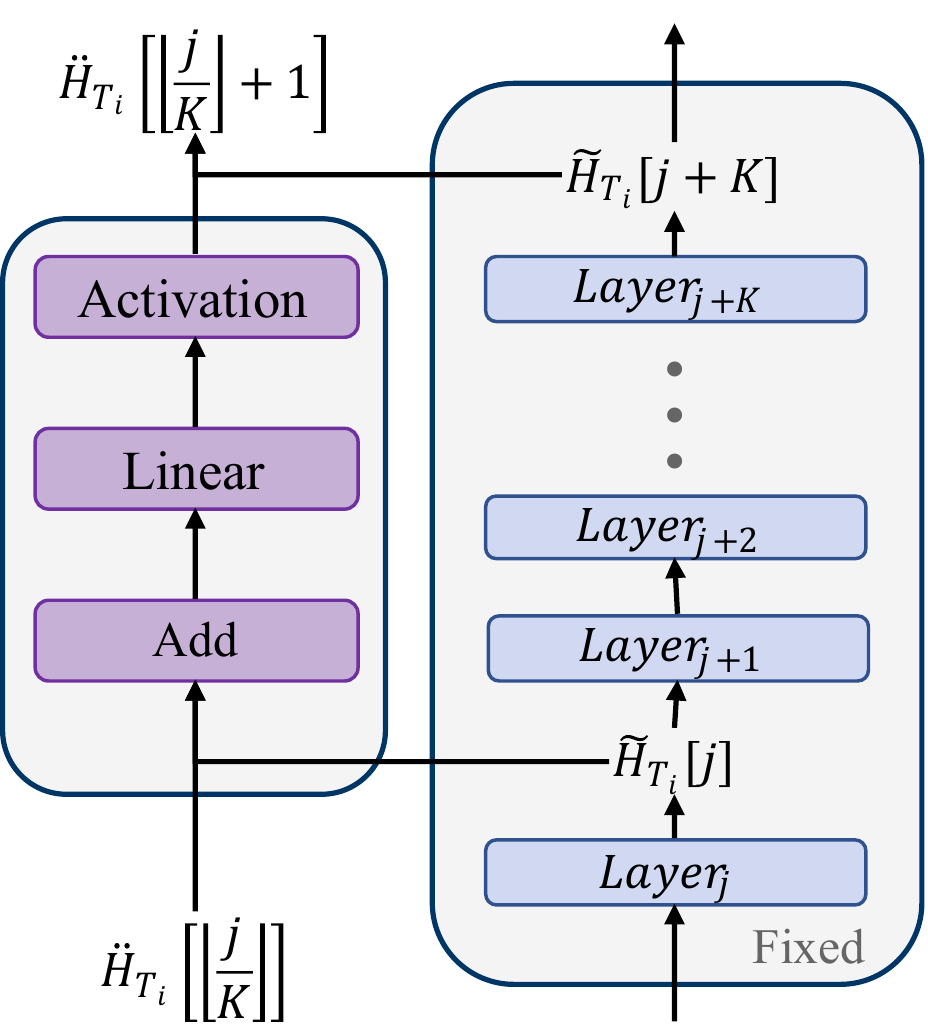}
    \caption{The processing of one unit of our \textbf{Protum-Res}. $\hat{H}_{T_i}[j]$ represents the hidden states after pooling of the $j_{th}$ layer in PLMs. $K$ is the number of the hidden layers cross the structure of PLMs and we treat it as a hyperparameter to decide the scale of updating parameters in our model. $\Ddot{H}_{T_i}\left[ \left \lfloor \frac{j}{K} \right \rfloor\right]$ is the output of last unit. }
    \label{fig:res_item}
\end{figure}

Firstly, we obtain $\hat{H}_{T_i}$ from the PLMs with fixed parameters as same as the step of \textbf{Protum-base} for avoiding the random combination between the multiple answer tokens. An instance of unit as is shown in the Figure \ref{fig:res_item}, in each unit we add $\hat{H}_{T_i}[j]$ to $\Ddot{H}_{T_i}\left[ \left \lfloor \frac{j}{K} \right \rfloor\right]$ which is calculated from the last unit. After that we bring the result of the sum input into linear layer in which the parameters are about to update during training. The output of this unit $\Ddot{H}_{T_i}\left[ \left \lfloor \frac{j}{K} \right \rfloor + 1\right]$ is obtained by activating with activation function.
\begin{equation}
    \Ddot{H}_{T_i}\left[ \left \lfloor \frac{j}{K} \right \rfloor + 1\right] = {\rm ReLU}({\rm FC}\left[\Ddot{H}_{T_i}\left[ \left \lfloor \frac{j}{K} \right \rfloor\right] + \hat{H}_{T_i}[j]\right])
\end{equation}
By setting the structural unit similar to the residual network, the hidden states of layers in the PLMs can be effectively utilized on the basis of the \textbf{Protum-base}. 
In other words, we can choose different values of $K$ which is actually the number of the hidden layers crossed by a single residual unit to make full use of almost all of the hidden layers.
Compared with the \textbf{Protum-base}, the scale of parameters of the fully connected layers can be handled by changing the value of $K$ and the number of unit so that the model can converge better in the various data sets. With such a simple residual unit design, we can solve the disadvantages of \textbf{Protum-base} well and get better performance than fine-tuning and many PT methods with the same scale of parameters when we fix the parameters of the PLMs. Last but not least, the residual units do not need to change the structure of the PLMs. That is to say it is very convenient when we save and load the weights of the residual units during tuning or predicting. For those downstream tasks, it is not necessary to copy the PLMs every time but only to load the weights of the units. This is one of the most significant advantages of the PT methods.
\section{Experiments \& Analysis}
In this section, we verify the performance of the above two models \textbf{Protum-base} and \textbf{Protum-Res} on three different scales of data size of sentence matching tasks on SuperGLUE \cite{wang2019superglue}. And we also introduce the details and steps about how we achieve our experiments. Lastly, an analysis about the result of our experiments will be given.
\subsection{NLU Tasks}
We choose CB, RTE and BoolQ from SuperGLUE \cite{wang2019superglue} to figure out the performance of our models in different scales of data sets. What these three types of data sets have in common is that all of them are typical sentence-pair matching tasks of NLP. As shown in the Table \ref{tab:dataset}, we give the detail about data size of three kinds of data sets. We also make the statistics about the average length of the sentence-pair tokenized by official tokenizer of the PLMs of these data sets which already have been added with the hard prompt tokens . 
\begin{table}[h]
    \centering
    \begin{tabular}{cccccc}
        \hline
         Dataset & $n_{class}$ & $|{\rm Train}|$ & $|{\rm Dev}|$ & Avg Length \\
        \hline
         CB & 3 & 250 & 57 & 92\\
         RTE & 2 & 2500 & 278 & 78\\
         BoolQ & 2 & 9427 & 3270 & 147\\
        \hline
    \end{tabular}
    \caption{The detail about data size of the data sets using to train our models.}
    \label{tab:dataset}
\end{table}
It is easy to find the difference between the scale of data size of these three data sets. The leap from hundreds of data to nearly ten thousands of data leads into the challenge of the robustness of our models. Therefore, our choice of these three types of data sets is strongly reasonable and also persuasive for us to verify the effectiveness of our models.
\subsection{Details of Experiments}
In this section, we present the experimental steps and details of the \textbf{Protum-base} and \textbf{Protum-Res} on the above data sets.
\subsubsection{Pre-training on Across Tasks}

In the process of PT, all of the parameters of the PLMs will be fixed to solve the problem of spatial redundancy caused by reduplication of the PLMs during FT. The PLMs have never seen those texts which are added by the hard prompt tokens or soft prompt tokens that we construct for a specific downstream task. Because we have not continually pre-trained the MLM on the texts of the task for understanding the semanteme of the words in the sentences.
For any given downstream tasks of NLP, people are going to comprehend the meaning of the texts first and then figure out the labels. 
There are two unreasonable issues if we take the original PLMs into consideration directly for PT. One is that obviously not in line with the way of people's thought, and the other is that the performance of the various PT methods are not satisfied as well as the FT methods. Additionally, \cite{gu2021ppt} also have proved that the effect of continuous pre-training will obtain much better performance. In this case, we pre-train the PLMs on the texts of across tasks first. 

We construct $T_i$ for three types of data sets respectively by add the appropriate hard prompt tokens as same as P-tuning \cite{liu2021gpt}. The $T_i$ is shown in the Table \ref{tab:template}. The approach of template construction is actually the same with different prompt tokens and answer tokens for different tasks. The only difference between training data and validation data is that we add the answer tokens in the training data while we remove them in the validation set to avoid the issue of labels leaking during the pre-training. Then all of the $T_i$ are sent to pre-train MLM after masking those tokens randomly selected. There are two essential strategies as follows for pre-training.
\begin{itemize}
    \item We take steps of dynamic masking \cite{liu2019roberta} and ten times duplication of data sets for pre-training.
    \item All of the selected tokens are masked during randomly masking.
\end{itemize}
The answer tokens also may be masked with a certain probability when we randomly select some tokens to mask. We have a conclusion that both the semanteme of texts and the logical relationship between the texts and the labels can be obtained in the way of pre-training. It is also the second purpose for us to pre-train to maximize the capability of the PLMs to comprehend the inference from the texts to the labels. About $15\%$ tokens of the sequence will be selected during pre-training. Only $80\%$ tokens are going to be masked while the left $20\%$ will keep original tokens or replace with the tokens randomly selected from vocabulary. We think both keeping original tokens and random replacement may reduce the performance of the models we proposed because what we actually do is classify the hidden states of the masked tokens. For this reason, all of the selected tokens are masked in the process of pre-training. The experiments also prove the effectiveness of our approach.
\begin{table*}[htbp]\small
    \centering
    \begin{tabular}{l|l}
    \toprule
    \toprule
    
    Dataset  & {\centering$T_i$ }\\
    \hline
     CB & [CLS] + Premise + [SEP] + Hypothesis + ['?', 'The', 'Answer', ':'] + Answer Tokens + ['.'] + [SEP]\\
     \hline
     RTE & [CLS] + Premise + ['Question', ':'] + Hypothesis + ['?', 'The', 'Answer', ':'] + Answer Tokens + ['.'] + [SEP]\\
     \hline
     BoolQ & [CLS] + Passage + ['The', 'Question', ':'] + Question + ['?', 'The', 'Answer', ':'] + Answer Tokens + ['.'] + [SEP]\\
     
    \bottomrule
    \bottomrule
    \end{tabular}
    \caption{Unifying Formats for the training dataset of three sentence-pair tasks. The formats of validation dataset is same as the training dataset only without the answer tokens.}
    \label{tab:template}
\end{table*}

\subsubsection{Hyperparameters}
We choose BERT-base and BERT-large as our baseline to verify the performance of our proposed two approaches on both large scale model and small model respectively. We set the max sequence length of the PLMs to $256$ during pre-training through the analysis of the average length of three types of data sets. Besides, we set the probability of randomly masked tokens for CB, RTE and BoolQ to $0.2$, $0.25$ and $0.2$ respectively. The learning rate for both \textbf{Protum-base} and \textbf{Protum-Res} are set to 1e-4, 2e-4, 1e-3 and 2e-3. In order to compare with other FT and PT methods, the size of the full connection layer is set to the hidden size of the PLMs. Lastly, the value of $K$, we choose the factors of the number of the hidden layers in the PLMs, which is $1,2,3,4,6,12$. 

\subsection{Main Results}
First of all, we pre-train both BERT-base and BERT-large on the data of the across tasks with actually the same strategies and hyperparameters. For different scale PLMs after pre-training on those data sets, we design various experiments for researching the performance of \textbf{Protum-base} and \textbf{Protum-Res} on three kinds of data sets.
\subsubsection{Protum-base}
We sent $\tilde{H}_{T_i}[j]$ to the fully connected layer to make classification in the process of \textbf{Protum-base}. Generally, we used to take the sequence output $\tilde{H}_{T_i}[-1]$ from the PLMs. However, we do not declare which the specific hidden layer could obtain the best result than other layers. On this condition, we choose RTE dataset to find out the performance of each layer in the PLMs.
Initially, only a single layer in the PLMs has been taken to compare its performance with each other. Meanwhile, we also bring the last few layers to make max or average pool and then send the output into classifier. The performance on RTE dataset is showing in Tabel \ref{tab:protum_base_1}. Lastly, according to the results, we have come to the following two conclusions in this part of experiments.
\begin{itemize}
    \item Generally, the hidden layer is closer to the input to get worse results.
    \item The third hidden layer always obtains the best performance both in large scale and small scale models on the almost all of the data sets.
\end{itemize}
\begin{table}[htp]
    \centering
    \begin{tabular}{lcc}
    \toprule
    \toprule
    
    Layer& \begin{tabular}[c]{@{}c@{}}BERT-base\\ (Acc.)\end{tabular} & \begin{tabular}[c]{@{}c@{}}BERT-large\\ (Acc.)\end{tabular} \\
    \hline
    $\tilde{H}_{T_i}[-1]$ & 61.2      & 71.8       \\
    $\tilde{H}_{T_i}[-2]$ & 63.5      & 74.1       \\
    $\tilde{H}_{T_i}[-3]$ & \underline{\textbf{67.2}}      & \underline{\textbf{75.8}}       \\
    $\tilde{H}_{T_i}[-4]$ & 65.0      & 71.5       \\
    MAX            & 64.3       & 74.0      \\
    AVG            & 63.2       & 73.7       \\
    
    \bottomrule
    \bottomrule
    \end{tabular}
    \caption{The results of another layers are not worthy to present in this table because of their worse results. We also run our code in the CB and BoolQ datasets and the performances which we do not show here also prove the predictable conclusion though only few times we can not obtain the best results using $\tilde{H}_{T_i}[-3]$ on the CB because of the fluctuation caused by the small data size of the validation data. The result obtained by max pooling and average pooling the last four layers.}
    \label{tab:protum_base_1}
\end{table}
It is easy to find that $\tilde{H}_{T_i}[-3]$ always achieve the better performance on the validation data than others both BERT-base and BERT-large. By the way, the result shows pooling the last few layers is not working to make improvement. 
In conclusion, the last third layer in the PLMs can obtain the best performance so that the result of the follow experiments attains by $\tilde{H}_{T_i}[-3]$. The results obtained by training the PLMs with three data sets is shown in the Table \ref{tab:protum_base_2}.
\begin{table*}[h]
    \centering
    \begin{tabular}{lcccccc}
    \toprule
    \toprule
    
    Dataset& \multicolumn{2}{c}{\begin{tabular}[c]{@{}c@{}}CB\\      (Acc.)\end{tabular}} & \multicolumn{2}{c}{\begin{tabular}[c]{@{}c@{}}RTE\\      (Acc.)\end{tabular}} & \multicolumn{2}{c}{\begin{tabular}[c]{@{}c@{}}BoolQ\\      (Acc.)\end{tabular}} \\
    \hline
    Model   & BERT-base & BERT -large   & BERT-base & BERT -large   & BERT-base & BERT -large\\
    \hline
    FT*     & 85.1  & 83.7  & 68.4  & 75.8  & 72.9  & 77.7\\
    FT      & none  & 83.6  & none  & 71.6  & none  & 77.4\\
    P-tuning*& 89.2  & 97.4  & 71.1  & 75.5  & 73.9  & 77.8\\
    P-tuning v2& none  & 94.6  & none  & 78.3  & none  & 75.8\\
    \textbf{Protum-base} & 89.2      & 91.2  & 67.9 & 75.8   & 71.8  & 73.9\\ 
    \textbf{Protum-Res} & \textbf{\underline{89.2}} & \textbf{\underline{94.6}}  & \textbf{\underline{72.1}} & \textbf{\underline{76.9}}   & \textbf{\underline{72.4}}  & \textbf{\underline{75.8}}\\ 
    
    \bottomrule
    \bottomrule       
    \end{tabular}
    \caption{The table shows the whole results for three data sets training on the PLMs. We report the same results of FT* and P-tuning* taken from \cite{gu2021ppt}. Additionally, the results of P-tuning v2 and FT are also obtained from \cite{liu2021pv2} and \cite{wang2019superglue} respectively.}
    \label{tab:protum_base_2}
\end{table*}

First of all, whether it is a large scale or small scale PLM, our \textbf{Protum-base} can basically perform decently as well as the results obtained by full-scale fine-tuning on these tasks. Although we have pre-trained the model before prompt tuning which is a kind of reasonable and effective measure, it is also sufficient to illustrate the effectiveness of our method. 
Secondly, the results are obviously affected by initialization and hyperparameters. The results obtained on the three types of validation datasets indicate the unstable factors exist in the process of training especially when the data size is quite small. There is no doubt that only the hidden states of the third hidden layer is utilized in \textbf{Protum-base} and what worse it is only the parameters of the classifier connected behind the PLMs that can be updated with the fixed PLMs. In our opinion, these are the main reason for the fluctuation of the results. Therefore, the performance of the \textbf{Protum-base} is not outstanding comparing with the other PT methods. In this case, we propose \textbf{Protum-Res} based on \textbf{Protum-base}.
\subsubsection{Protum-Res}
\begin{figure*}[ht]
	\centering
		\begin{minipage}{0.4\textwidth}
			\centering
			\begin{tabular}{c|cc}
            \toprule
            \toprule
            
            \multirow{2}*{K} & \multicolumn{2}{c}{\begin{tabular}[c]{@{}c@{}}RTE\\      (Acc.)\end{tabular}} \\
            \hline
             & BERT-large & BERT-base \\
            \hline
            1 & 74.4 & 66.4 \\
            2 & 75.1 & 67.2 \\
            3 & \textbf{\underline{75.5}} & \textbf{\underline{67.9}} \\
            4 & 75.4 & 63.9 \\
            6 & 73.7 & 63.2 \\
            12 & 72.0 & 62.5\\
            \bottomrule
            \bottomrule
            \end{tabular}
            \tabcaption{The accuracy of the validation dataset of RTE when we set different value to $K$ during training \textbf{Protum-Res}. The results show the value around the median of $K$ can obtain better performance.}
            \label{tab:k_value}
		\end{minipage}
	\hspace{0.5in}
		\begin{minipage}[h]{0.45\linewidth}
		    \centering
            \includegraphics[scale=0.45]{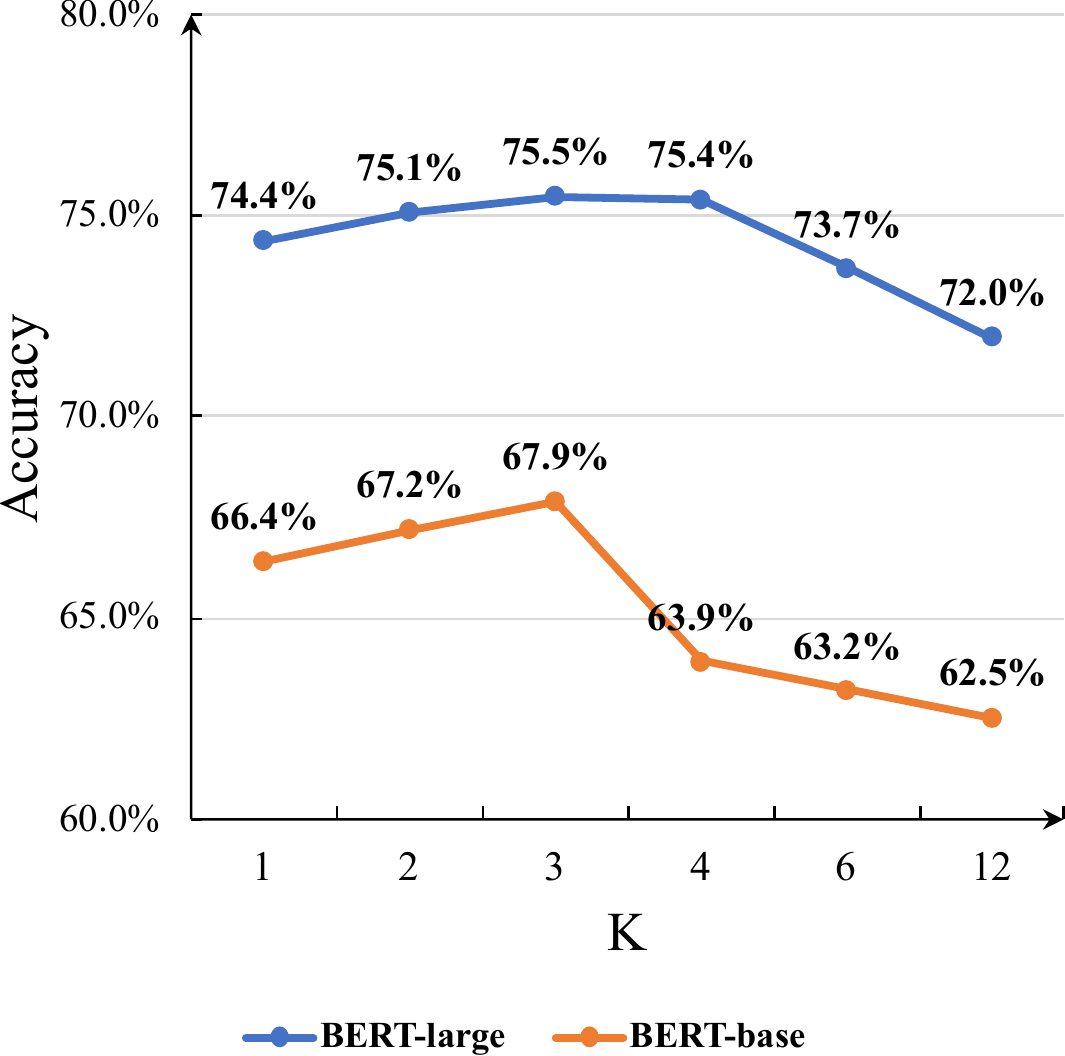}
            \figcaption{This figure also indicates there still exists a huge gap between the small language models and the large language models.}
            \label{fig:k_value}
		\end{minipage}%
\end{figure*}

Owing to the additional residual blocks inserted into the PLMs, theoretically our \textbf{Protum-Res} can converge faster that \textbf{Protum-base} and the better performances is about to be obtained with making full usage of the hidden states and higher amount of parameters to update. For the purpose that how the units affect the performance of the models, we design experiments to make full research of them. $K$, which equals the number of the hidden layer that one residual unit crosses, it is a crucial factor for \textbf{Protum-Res} while it decides the amount of the units inserting into the PLMs. Simultaneously, it is also has a effect on the updating parameters of \textbf{Protum-Res} due to the fully connected layer in the each unit. There is no denying that it is necessary for us to research $K$. Initially, we design the experiments on RTE dataset. The results of this part is showing at Table \ref{tab:k_value} and Figure \ref{fig:k_value}.

Generally, the less parameters of the \textbf{Protum-Res}, the worse accuracy of validation dataset will be. As expected, the performances are little satisfied even worse than \textbf{Protum-base} when we set $K$ to 6 or 12.
The results are exactly different with what we have in our opinions, which present an unexpected phenomenon that the smaller $K$ that means the maximum of the residual units in the PLMs acquires worse performances on the validation set of RTE. That is to say by adding a amount of the residual units specially close to $N$ can not bring decent improvements for our model.
At the same time, we also obtain this conclusion when we do the same experiments on the CB and BoolQ datasets.

According to the conclusion in \textbf{Protum-base} that the hidden states closer to the input layer contain less useful information for any downstream tasks leads into the terrible performances. Nevertheless, the residual unit always has been added to the input layer of the PLMs in the above experiments. There is no doubt that the hidden states of the front layers reduce the performance and robustness of our model. For this reason, we set a extra hyperparameter $S$ which is the position where the first residual unit started. $S\in \left[1, \frac{N}{K} \right]$. The smaller the value of $S$, the closer the first residual to the input layer. In this case, we select the best $K=3$ to design the experiments on RTE dataset for exploring the influence of $S$. As it is shown in the Figure \cite{fig:s_value}.
\begin{figure}
    \centering
    \includegraphics[scale=0.4]{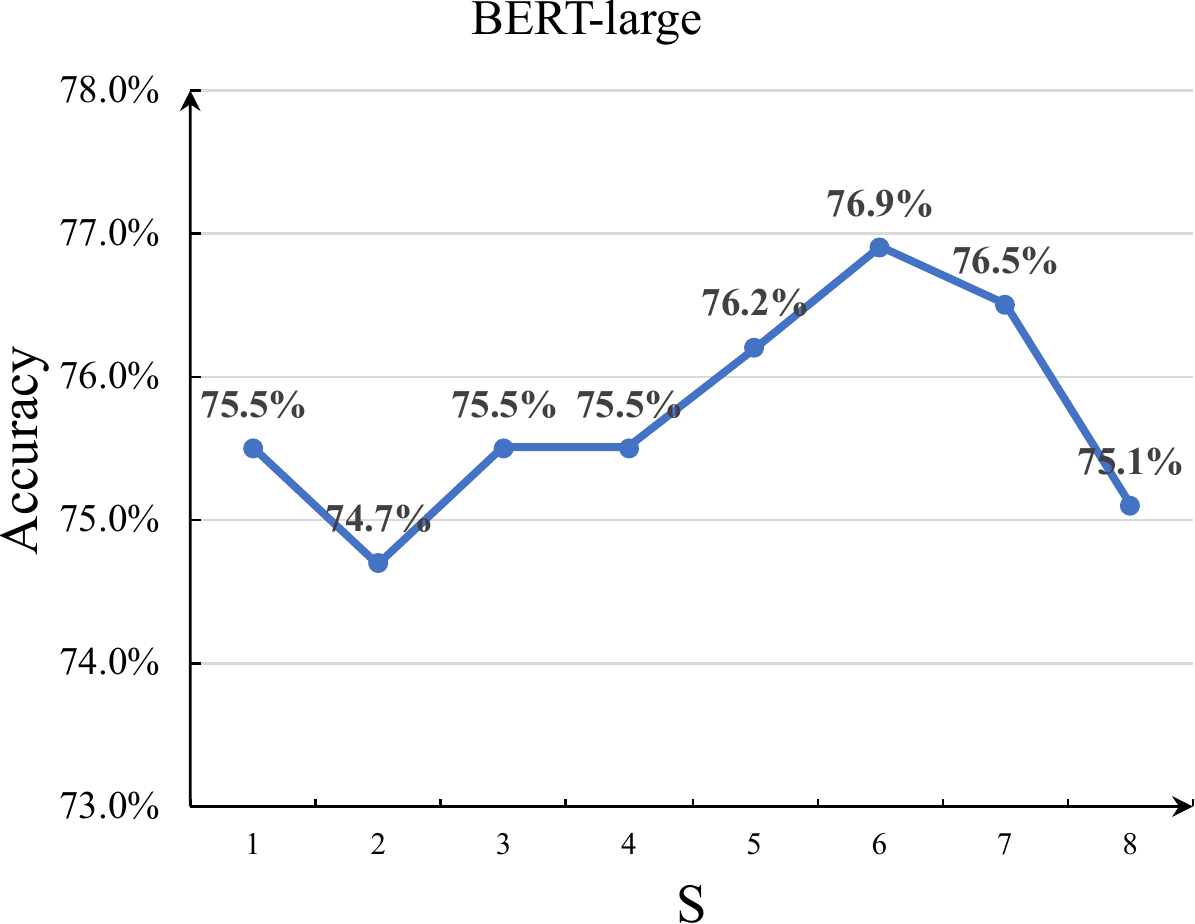}
    \caption{The accuracy of the validation set of RTE when the different values of $S$ are selected during training. The best result obtained by setting the $S$ to 6.}
    \label{fig:s_value}
\end{figure}
According to the experimental results, we first draw the conclusions:
\begin{itemize}
    \item The $S$ has a great influence on the performance of \textbf{Protum-Res}.
    \item The $S$ is too large or small to obtain better results.
\end{itemize}
The value of $K$ decides the number of the residual units and the position of the first unit. Not only does it affect the scale of parameters which can be updated but also which part of the hidden layers can be used during training. 
The smaller the value of $S$, the more the hidden layers closed to the input layer are input into the network. Although there are a great amount of parameters, it is still difficult to obtain decent results. 
In contrast, the smaller the number of residual units results in less use of the hidden states when the value of $S$ gets larger.
Therefore, in order to trade off the above two effects, selecting the intermediate value is appropriate to obtain better results.

Finally, we compare our method with other recently advanced PT methods on these three types of data sets. The results are showing in the Table \ref{tab:protum_base_2}. The results obtained by synthesizing the above two experimental conclusions while we select the proper $K$ and $S$ to get the best results on these three types of data sets. It is easy to find that the results of our methods whether in both BERT-base and BERT-large models of the validation dataset of the CB and RTE greatly surpass the results obtained by the FT which report at SuperGLUE \cite{wang2019superglue}. 
On the validation dataset of BoolQ, there is still a certain gap with fine-tuning method. However, comparing against other PT methods, we can get the same level results in the case of the small parameters of our model than others. In our \textbf{Protum}, we choose to discard the verbalizer and directly use the hidden states of the PLMs for classification. Through inquiry and comparative experiments, whether in large language models or small language models and large scale data sets or small scale data sets, our model has obtained excellent results under the condition of freezing the parameters of the PLMs. Finally, the experimental results show that our proposed models are effective and meaningful.
\section{Conclusion}
In this paper, we first propose \textbf{Protum-base} to solve some issues of the PT methods we raised in Chapter 1. The hidden states of the answer tokens are directly used to make classification instead of being converted into a specific natural language word for avoiding random composition of the answer tokens and reducing the disadvantage of weakening the performance of the model owing to the alternative choices of the answer tokens. 
Through the analysis and experiments of the \textbf{Protum-base}, aiming at its shortcomings, we propose \textbf{Protum-Res} by adding the residual units which is the enhancement model based on the \textbf{Protum-base}. By studying the influence of various factors of the residual unit, we finally draw a reasonable conclusion and get decent results on multiple data sets with different data sizes. Especially for the small language models which can not obtain considerable results using other PT methods before, our models give play to their potential and get better performance and robustness. 
Meanwhile, the fast training speed and without any duplication of the PLMs for each downstream task, our methods further promote the deployment of large language models in practical applications. 
In the next step, we prepare to explore the robustness of our \textbf{Protum} when the different hard prompts are added to the sentences. Last but not least, the residual unit are not input into the hidden layers of the PLMs. What we actually do is that using the hidden states sufficiently by the residual unit make classification. The further research of the residual unit will also be included in our plan.

\nocite{langley00}

\bibliography{example_paper}
\bibliographystyle{icml2022}



\end{document}